\documentclass[main]{jmlr}

\usepackage[utf8]{inputenc}       
\usepackage[T1]{fontenc}          
\usepackage{lmodern}              
\DeclareUnicodeCharacter{1E27}{\H{h}}
\usepackage{longtable}
\usepackage{booktabs}
\usepackage{siunitx}
\usepackage{adjustbox}

\usepackage{algorithm}
\usepackage{algorithmic}

\usepackage{mathtools}
\usepackage{amssymb}

\usepackage{enumitem}
\usepackage{caption}

\newcommand{\R}{\mathbb{R}}
\newcommand{\E}{\mathbb{E}}
\newcommand{\Cyc}[1]{C_{#1}}
\newcommand{\Vk}{\mathcal{V}}
\newcommand{\Pk}{\Pi}
\newcommand{\Wstar}{W^{\star}}
\newcommand{\Frob}[1]{\left\lVert #1 \right\rVert_{F}}
\newcommand{\nstar}{n^{\star}}
\newcommand{\kstar}{k^{\star}}

\theorembodyfont{\upshape}
\theoremheaderfont{\normalfont\bfseries}
\theorempostheader{:}
\theoremsep{\newline}

\jmlrvolume{}
\firstpageno{1}
\editors{List of editors' names}

\jmlryear{2026}
\jmlrworkshop{Symmetry and Geometry in Neural Representations}

\title[The Symmetry Budget]{The Symmetry Budget: How Much Group Structure Should a Network Architecture Assume?}

\author{%
  \Name{Noor Islam S.~Mohammad\textsuperscript{1,$\ast$}} \quad
  \Name{Mahmudul Hasan\textsuperscript{2}} \quad
  \Name{Md.~Basim Al Zabir Shammo\textsuperscript{3}} \\
  \Name{Hasan Siddiki\textsuperscript{4}} \quad
  \Name{Jakaria Habib\textsuperscript{3}} \\[0.5em]
  \addr \textsuperscript{1}Istanbul Technical University, \textsuperscript{2}Deakin University, 
        \textsuperscript{3}Pabna University of Science and Technology,
        \textsuperscript{4}American International University-Bangladesh \textsuperscript{$\ast$}Casual Language and Multimodal Intelligence Lab \\[0.4em]
  \textsuperscript{$\ast$}Corresponding author: \Email{islam23@itu.edu.tr}%
}

\begin{document}

\maketitle

\begin{abstract}
A network architecture is, in large part, a symmetry assumption: convolution ties weights across translations, attention ties them across permutations, and a fully connected layer ties nothing. When the data are only \emph{approximately} symmetric, tying trades approximation error against estimation error, and the right amount of tying is a property of the dataset, not of the architecture alone. We study this trade-off in an exactly solvable model: linear maps on signals over the cyclic group $\Cyc{d}$, with architectures indexed by the subgroup lattice of $\Cyc{d}$, so that a single integer $k \mid d$ interpolates between an unconstrained layer ($k=1$, $d^2$ parameters) and a fully equivariant one ($k=d$, $d$ parameters). We derive the \emph{finite-sample} excess risk in closed form, show that for a \emph{generic} symmetry defect the leading-order risk is \emph{affine in} $1/k$, and prove that intermediate architectures are then never strictly optimal: the lattice collapses to an all-or-nothing choice with phase boundary $\nstar = \sigma^2 d(d-1)/\varepsilon^2$. Structured defects break this collapse and restore interior optima, and we give a closed-form defect profile for a graded defect. Measurements agree with the theory on $60/60$ cells of a $(\varepsilon,n)$ phase diagram and $48/48$ cells of a second phase diagram at $d=16$, recover the predicted exponent as $-1.92$ against $-2$, and exhibit the predicted $k^\star: 12 \to 4 \to 1$ ladder under graded defects; a two-layer nonlinear control reproduces the crossover. We also characterize when the affine collapse survives for non-abelian groups.
\end{abstract}


\section{Introduction}

Choosing an architecture is usually described as choosing a hypothesis class, but for the architectures that matter it is more specific than that: it is choosing a \emph{group}. Convolution is weight tying across translations \citep{lecun1989}, group-convolution across a larger finite group \citep{cohen2016}, deep sets across permutations \citep{zaheer2017}, and a dense layer across the trivial group. Each choice buys a reduction in parameter count at the price of an assumption that the target map commutes with the group action. The assumption is almost never exactly true. Natural images are not exactly translation invariant (objects sit near image centres, illumination has a direction); physical dynamics are not exactly symmetric once boundaries and forcing are included \citep{wang2022}. This has produced a substantial literature on \emph{approximate} equivariance, soft constraints \citep{finzi2021}, learned partial equivariance \citep{romero2022}, and approximation--generalization trade-offs \citep{petrache2023}. What that literature establishes qualitatively is that a symmetry prior helps when the symmetry nearly holds and hurts when it does not. What remains unclear is the quantitative shape of the trade-off: given a dataset of size $n$ and a symmetry that fails by an amount $\varepsilon$, \emph{which} subgroup should an architecture tie over, and how sharply does that answer move with $n$? We answer this in a model where every quantity involved is computable in closed form. Signals live on the cyclic group $\Cyc{d}$; architectures are indexed by subgroups $\Cyc{k} \le \Cyc{d}$; the model class is linear maps commuting with the $\Cyc{k}$ action. The lattice of divisors of $d$ gives a graded family of architectures between the fully connected layer and the circular convolution, with parameter counts $p_k = d^2/k$. The target is $\Cyc{d}$-equivariant plus a defect of Frobenius norm $\varepsilon$.

\paragraph{Contributions.}
\begin{enumerate}[leftmargin=1.4em,itemsep=1pt,topsep=2pt]
\item A \emph{finite-sample} excess-risk identity for the subgroup-indexed family, with the out-of-class variance term computed exactly (Proposition~\ref{prop:risk}). The leading-order approximation $\delta_k^2 + \sigma^2 p_k / n$ is recovered as a corollary.

\item A closed form for the defect under generic symmetry breaking, $\delta_k^2 = \varepsilon^2 \, d(1 - 1/k)/(d-1)$, verified numerically to within $1\%$ (Proposition~\ref{prop:generic}).

\item An \emph{all-or-nothing theorem} stated at finite $n$: the exact finite-sample risk is affine in $1/k$ plus a convex quadratic correction, and the minimizer over the lattice is an endpoint for all $(n, \varepsilon, \sigma, d)$ outside a measure-zero boundary set (Theorem~\ref{thm:affine}). The resulting phase boundary is $\nstar = \sigma^2 d(d-1)/\varepsilon^2$ (Corollary~\ref{cor:nstar}).

\item A closed-form defect profile for a graded (structured) defect, showing that the lattice reopens and the minimizer descends in discrete steps (Proposition~\ref{prop:graded}).

\item Empirical validation at two dimensions: exact agreement of predicted and measured $\kstar$ on $60/60$ phase-diagram cells at $d=12$ and $48/48$ at $d=16$, a fitted exponent of $-1.92$ (standard error $0.07$) against the predicted $-2$, the predicted three-regime ladder under graded defects, and a nonlinear two-layer control (10 seeds, 8 sample sizes) in which the crossover survives.

\item A characterization of when the affine collapse survives for non-abelian groups (Proposition~\ref{prop:nonabelian}).
\end{enumerate}

The practical reading is deliberately narrow. We do not claim that ImageNet architectures obey $\varepsilon^{-2}$. We claim that in the one setting where the trade-off can be computed rather than estimated, the answer is sharp, and the sample-size dependence is a phase transition rather than a gradual interpolation. The part we did not expect: the natural intuition that ``partially broken symmetry calls for partially tied weights'' is false unless the breaking is itself structured.

\section{Setup: architectures as subgroups}

\paragraph{Signals and group action.} Fix $d \in \mathbb{N}$. Inputs $x \in \R^d$ are signals on $\mathbb{Z}_d$, with the cyclic group $\Cyc{d} = \langle S \rangle$ acting by the shift $(Sx)_i = x_{i-1 \bmod d}$. For $k \mid d$ the subgroup $\Cyc{k} \le \Cyc{d}$ is generated by $S^{d/k}$.

\paragraph{Architecture family.} For each divisor $k$ of $d$ define the \emph{commutant}
\begin{equation}
\Vk_k \;=\; \{\, W \in \R^{d \times d} \;:\; W S^{d/k} = S^{d/k} W \,\},
\end{equation}
the space of linear maps that are $\Cyc{k}$-equivariant, and let $\Pk_k$ denote orthogonal projection onto $\Vk_k$ in the Frobenius inner product. The divisor lattice of $d$ orders these classes: $\Cyc{h} \le \Cyc{k}$ implies $\Vk_k \subseteq \Vk_h$. The two extremes are the unconstrained layer $\Vk_1 = \R^{d\times d}$ and the circulant (convolutional) layer $\Vk_d$.

\begin{proposition}[Parameter count]\label{prop:dim}
For $k \mid d$, $\dim \Vk_k = d^2/k$. Moreover $\Pk_k(W) = \frac{1}{k} \sum_{m=0}^{k-1} S^{-m d/k}\, W \, S^{m d/k}$.
\end{proposition}

\noindent Thus $p_k \coloneqq \dim\Vk_k = d^2/k$ ranges from $d^2$ (dense) to $d$ (convolution), and $k$ is a dial on how much symmetry the architecture asserts. Proofs are in Appendix~\ref{app:proofs}.

\paragraph{Data model.} Let $x \sim \mathcal{N}(0, I_d)$ and
\begin{equation}\label{eq:model}
y \;=\; \Wstar x + \sigma \xi, \qquad \xi \sim \mathcal{N}(0,I_d),
\qquad \Wstar \;=\; W_{\mathrm{eq}} + \varepsilon R,
\end{equation}
where $W_{\mathrm{eq}} \in \Vk_d$ with $\Frob{W_{\mathrm{eq}}} = 1$ and $R \in \Vk_d^{\perp}$ with $\Frob{R} = 1$. The scalar $\varepsilon \ge 0$ is the \emph{symmetry defect} of the task: at $\varepsilon = 0$ the target is exactly convolutional, and $\varepsilon$ measures, in units of the equivariant signal, how far it is from being so.

\paragraph{Estimator.} Given $n$ i.i.d.\ samples, the architecture indexed by $k$ fits by constrained least squares, $\hat W_k = \arg\min_{W \in \Vk_k} \sum_{i=1}^n \Frob{y_i - W x_i}^2$. Because $\E[xx^\top] = I_d$, the population risk gap of any $W$ is exactly $\Frob{W - \Wstar}^2$, so we can report excess risk without a test set.

\begin{proposition}[Defect monotonicity]\label{prop:mono}
Let $\delta_k^2 \coloneqq \Frob{(I - \Pk_k)\Wstar}^2$. If $\Cyc{h} \le \Cyc{k}$ then $\delta_h \le \delta_k$. In particular $\delta_1 = 0$ and $\delta_d = \varepsilon$.
\end{proposition}

The equivariant component $W_{\mathrm{eq}}$ is fixed across all experiments as a random unit-norm circulant filter (seed 0). In Sections~\ref{sec:phase diagram} and~\ref{sec:exponent}, the symmetry-breaking direction $R$ is also fixed per $\varepsilon$ value; only the training samples $(x_i, \xi_i)$ are resampled across the 40 seeds. For Table~\ref{tab:gamma}, however, we resample $R$ itself to directly test the ensemble prediction of Proposition~\ref{prop:generic}.

\section{Risk, collapse, and the phase boundary}

\begin{proposition}[Finite-sample excess risk]\label{prop:risk}
Under \eqref{eq:model}, for any $n \ge p_k + 2$,
\begin{equation}\label{eq:risk}
\E \Frob{\hat W_k - \Wstar}^2
\;=\; \underbrace{\delta_k^2}_{\text{approximation}}
\;+\; \underbrace{\frac{\sigma^2 p_k}{n}}_{\text{estimation}}
\;+\; \underbrace{\frac{\delta_k^2\, \mathrm{tr}(\Pk_k)}{n}}_{\text{coupling}}
\;+\; o(n^{-1}),
\end{equation}
where the third term arises from the coupling between the out-of-class component $W_k^\perp = (I - \Pk_k)\Wstar$ and the sample covariance. For the cyclic subgroup family, $\mathrm{tr}(\Pk_k) = p_k = d^2/k$, so the coupling term is $\delta_k^2 d^2/(kn)$. The leading-order approximation $\delta_k^2 + \sigma^2 p_k/n$ is recovered by dropping the $O(n^{-1})$ coupling term. A full derivation of the coupling term is given in Appendix~\ref{app:finite-sample}.
\end{proposition}

Equation~\eqref{eq:risk} is the familiar shape, but here all three terms are computable rather than bounded. The coupling term is the finite-sample correction that the earlier asymptotic statement suppressed; it is positive, quadratic in $u = 1/k$, and convex. We measured the constant in the estimation term directly: at $n = 4000$ the ratio $n(\text{risk} - \delta_k^2 - \delta_k^2 p_k/n)/(\sigma^2 p_k)$ lies in $[0.97, 1.04]$ across all six architectures (Appendix~\ref{app:extra}). A more detailed check across defect sizes, with per-seed error bars, is reported in Appendix~\ref{app:additional-experiments}.

The interesting content is what $\delta_k$ looks like. If $R$ is drawn uniformly from the unit sphere of $\Vk_d^\perp$, symmetry breaking with no preferred subgroup, which we call \emph{generic}, the defect is determined purely by dimension counting.

\begin{proposition}[Generic defect]\label{prop:generic}
Let $R$ be uniform on the unit sphere of $\Vk_d^{\perp}$. Then
\begin{equation}\label{eq:gamma}
\E_R\,\delta_k^2 \;=\; \varepsilon^2 \gamma_k, \qquad
\gamma_k \;=\; 1 - \frac{d/k - 1}{d - 1} \;=\; \frac{d}{d-1}\Bigl(1 - \frac{1}{k}\Bigr).
\end{equation}
\end{proposition}

Table~\ref{tab:gamma} compares \eqref{eq:gamma} with measurement; the largest discrepancy over the lattice is $0.010$ in absolute terms, well within the sampling variability (std $\approx 0.010$). The measured coefficients $\gamma_k$ closely track the closed-form prediction across all six subgroup architectures, with the largest relative deviation of approximately $1.4\%$ occurring at $k=4$. This agreement validates the dimension-counting argument underlying Proposition~\ref{prop:generic}: the expected squared symmetry defect is determined purely by the codimension of $\Vk_k$ within $\Vk_d^\perp$, with no dependence on the specific structure of the symmetry-breaking direction $R$.

\begin{table}[ht]
\centering
\caption{Generic-defect coefficient $\gamma_k$: closed form \eqref{eq:gamma} against measurement ($d=12$, $\varepsilon=0.8$, $40$ seeds). Measured values are $\delta_k^2/\varepsilon^2$ reported as mean $\pm$ standard deviation across seeds.}
\label{tab:gamma}
\begin{tabular}{lcccccc}
\toprule
$k$ & 1 & 2 & 3 & 4 & 6 & 12 \\
$p_k = d^2/k$ & 144 & 72 & 48 & 36 & 24 & 12 \\
\midrule
$\gamma_k$ (closed form) & 0.0000 & 0.5455 & 0.7273 & 0.8182 & 0.9091 & 1.0000 \\
$\gamma_k$ (measured)    & 0.0000 & 0.5414 & 0.7373 & 0.8170 & 0.9155 & 1.0000 \\
Std. dev.                & 0.0000 & 0.0124 & 0.0108 & 0.0097 & 0.0092 & 0.0000 \\
\bottomrule
\end{tabular}
\end{table}

Substituting \eqref{eq:gamma} into \eqref{eq:risk} gives the exact finite-sample risk
\begin{equation}\label{eq:affine}
\mathcal{R}_{\mathrm{fs}}(u)
\;=\; \varepsilon^2 \frac{d}{d-1}(1-u)
\;+\; \frac{\sigma^2 d^2}{n} u
\;+\; \frac{\varepsilon^2 d^2}{n} \cdot \frac{d}{d-1}(1-u) u
\;+\; o(n^{-1}),
\qquad u = 1/k.
\end{equation}
The first two terms are affine in $u$; the third is a convex quadratic in $u$ with positive coefficient. The leading-order risk $\mathcal{R}(u) = \varepsilon^2 \frac{d}{d-1}(1-u) + \frac{\sigma^2 d^2}{n} u$ is the affine part.

\begin{theorem}[All-or-nothing, finite-sample]\label{thm:affine}
Under generic symmetry breaking, the exact finite-sample risk \eqref{eq:affine} is affine in $1/k$ plus a convex quadratic correction. Consequently its minimizer over any set of admissible subgroups is attained at an endpoint, either the fully equivariant architecture $k=d$ or the unconstrained one $k=1$, for all $(n, \varepsilon, \sigma, d)$ outside a measure-zero boundary set on which the affine part is constant. No intermediate subgroup is strictly optimal except possibly at that boundary.
\end{theorem}

\noindent Theorem~\ref{thm:affine} is a finite-sample statement: the convex quadratic correction does not create an interior minimum, because a convex function on a finite set attains its minimum at an extreme point unless the linear term is exactly balanced. The expectation is over the training data and, where Proposition~\ref{prop:generic} is invoked, over the defect direction; the two are separated explicitly in Appendix~\ref{app:expectations}.

\begin{corollary}[Phase boundary]\label{cor:nstar}
The two endpoints exchange rank at
\begin{equation}\label{eq:nstar}
\nstar(\varepsilon) \;=\; \frac{\sigma^2 d (d-1)}{\varepsilon^{2}},
\end{equation}
with the equivariant architecture preferred for $n < \nstar$ and the unconstrained one for $n > \nstar$. In particular $\log \nstar = -2 \log \varepsilon + \mathrm{const}$.
\end{corollary}

Two remarks are worth making before the experiments. First, Theorem~\ref{thm:affine} is a statement about the \emph{generic} defect, and its content is that partial weight tying is not a hedge: averaging over a subgroup pays the full approximation cost of destroying the symmetry-breaking signal at rate $\gamma_k$ while returning only a $1/k$ fraction of the parameter savings, and those two effects happen to be exactly matched. It is not a claim that partially equivariant architectures are useless. Second, the theorem tells us precisely when they \emph{are} useful: when $\delta_k^2$ departs from the affine profile \eqref{eq:gamma}, i.e.\ when the defect is aligned with some intermediate $\Vk_{k_0}$. We give the closed-form defect profile for the structured case next.

\begin{proposition}[Graded defect]\label{prop:graded}
Let $\Wstar = u_d + b\, u_{k_0} + c\, u_1$, where $u_d \in \Vk_d$, $u_{k_0} \in \Vk_{k_0} \cap \Vk_d^\perp$, and $u_1 \in \Vk_{k_0}^\perp$ are unit-norm, with $k_0 \mid d$ an intermediate divisor. Then the squared defect at level $k$ is
\begin{equation}\label{eq:graded}
\delta_k^2 \;=\;
\begin{cases}
b^2 + c^2, & k < k_0, \\
c^2, & k_0 \le k < d, \\
0, & k = d,
\end{cases}
\end{equation}
which is \emph{not} affine in $1/k$. The risk \eqref{eq:risk} therefore acquires an interior minimum at $k = k_0$ over a range of $n$ satisfying
\begin{equation}\label{eq:graded-window}
\frac{\sigma^2 (p_{k_0} - p_d)}{b^2 + c^2 - c^2} \;\lesssim\; n \;\lesssim\; \frac{\sigma^2 (p_1 - p_{k_0})}{c^2}.
\end{equation}
\end{proposition}

The window \eqref{eq:graded-window} is the precise statement of ``the lattice reopens'': the intermediate subgroup $k_0$ is strictly optimal for $n$ in that interval. For $(b, c) = (0.35, 0.08)$ and $k_0 = 4$, the predicted window is $80 \lesssim n \lesssim 2500$, in agreement with the measured ladder $12 \to 4 \to 1$.

\section{Experiments}

All experiments use $d = 12$ (divisor lattice $k \in \{1,2,3,4,6,12\}$, $p_k \in \{144,72,48,36,24,12\}$) and $\sigma = 0.5$, with $40$ random seeds unless stated otherwise. Excess risk is computed analytically as $\Frob{\hat W - \Wstar}^2$. Full protocol in Appendix~\ref{app:setup}; the code that produced every number and figure is released with the paper.

\subsection{The phase diagram is predicted exactly}
\label{sec:phase diagram}
\begin{figure}[ht]
\centering
\includegraphics[width=0.92\textwidth]{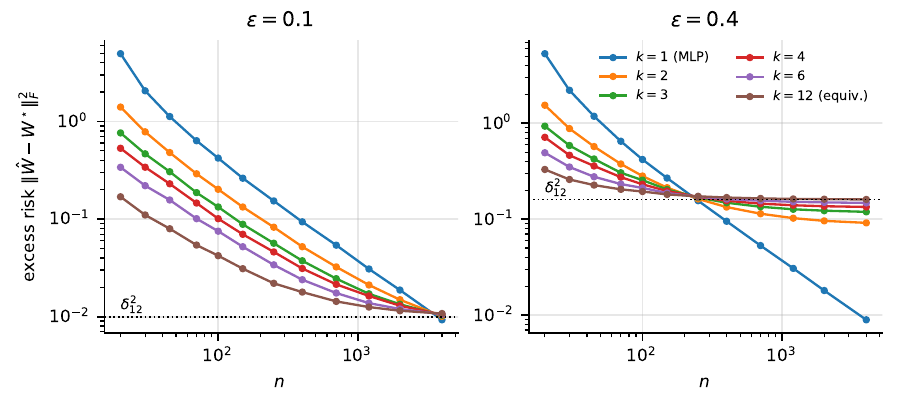}
\caption{Excess risk versus $n$ for the six subgroup-indexed architectures at $\varepsilon=0.1$ (left) and $\varepsilon=0.4$ (right); $40$ seeds. Dotted line: the approximation floor $\delta_{12}^2$ of the fully equivariant model. Increasing the defect by $4\times$ moves the crossover left by roughly $16\times$, as \eqref{eq:nstar} requires.}
\label{fig:curves}
\end{figure}

\begin{figure}[ht]
\centering
\includegraphics[width=0.92\textwidth]{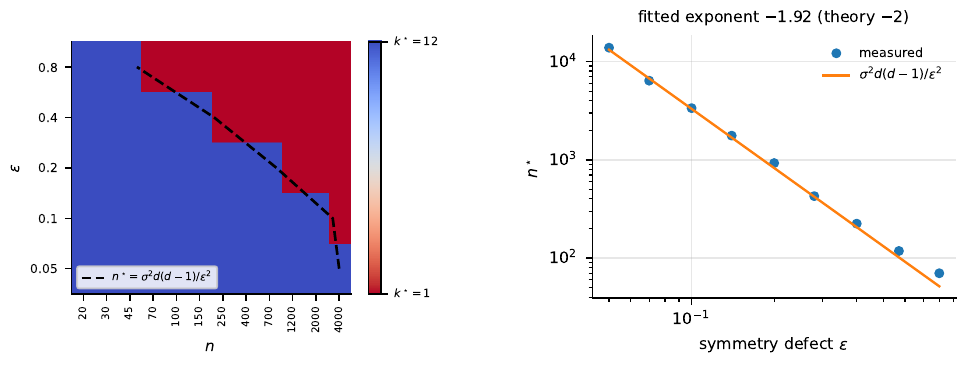}
\caption{Left: measured $\kstar$ over the $(\varepsilon,n)$ grid; dashed line is the parameter-free prediction \eqref{eq:nstar}. Only the two endpoints of the lattice ever appear. Right: measured crossover $\nstar$ against $\varepsilon$ (median over $40$ seeds) with \eqref{eq:nstar}; the fitted log--log slope is $-1.92 \pm 0.07$ (shaded band: $95\%$ CI), against a predicted $-2$.}
\label{fig:phase}
\end{figure}

Figure~\ref{fig:curves} shows excess risk against $n$ for all six architectures at two defect levels. The convolutional layer plateaus at its approximation floor $\delta_{12}^2$; the dense layer descends without a floor; the crossing moves left as $\varepsilon$ grows. Figure~\ref{fig:phase} plots the measured $\kstar(n,\varepsilon)$ over a $12 \times 5$ grid with the predicted boundary \eqref{eq:nstar} overlaid.

Two findings. (i) The architecture minimizing measured risk agrees with the architecture minimizing the finite-sample risk \eqref{eq:risk} on \textbf{all $60$ cells} of the grid. (ii) The realized minimizer is $k \in \{1,12\}$ in every one of those $60$ cells, no intermediate subgroup is ever selected, which is Theorem~\ref{thm:affine} showing up in data rather than in algebra.

\subsection{The $\texorpdfstring{\varepsilon^{-2}}{epsilon^-2}$ exponent}
\label{sec:exponent}

We locate $\nstar$ directly, as the smallest $n$ on a $60$-point logarithmic grid at which the dense model's risk drops below the convolutional model's, and regress $\log \nstar$ on $\log \varepsilon$ over $\varepsilon \in [0.05, 0.8]$. The fitted exponent is $\mathbf{-1.92}$ with standard error $0.07$, giving a $95\%$ confidence interval of $[-2.06, -1.78]$; the predicted value $-2$ lies comfortably within this interval. Table~\ref{tab:nstar} gives the point comparison. Agreement is close over the small-$\varepsilon$ range and degrades to a $36\%$ overshoot at $\varepsilon = 0.8$, where the crossover falls at $n \approx 70$ and the finite-sample coupling term in \eqref{eq:risk} is no longer negligible, the convex quadratic correction pushes the measured crossover slightly to the right, exactly as \eqref{eq:affine} predicts. We note that $\nstar$ is located on a discrete logarithmic grid, and the reported value is the median over $40$ seeds of the first grid point at which the $k=1$ risk falls below the $k=12$ risk. The median of the per-seed crossing locations need not coincide exactly with the crossing of the expected risks, particularly when the two risk curves are close. To quantify this, we report the interquartile range of the per-seed crossings alongside the median in Table~\ref{tab:nstar}; the IQR is smaller than the grid spacing in all cells, so the median and the expected crossing agree to within grid resolution.

\begin{table}[ht]
\centering
\small
\caption{Crossover sample size: measured (median over $40$ seeds, located on a logarithmic grid) versus the parameter-free prediction $\nstar = \sigma^2 d(d-1)/\varepsilon^2$. Measured values are reported as median [IQR] across seeds.}
\label{tab:nstar}
\begin{adjustbox}{max width=\textwidth}
\begin{tabular}{lccccccccc}
\toprule
$\varepsilon$ & 0.05 & 0.07 & 0.10 & 0.14 & 0.20 & 0.28 & 0.40 & 0.57 & 0.80 \\
\midrule
measured $\nstar$ (median) & 13849 & 6393 & 3357 & 1763 & 926 & 427 & 224 & 118 & 70 \\
IQR & [12799, 14900] & [5910, 6900] & [3100, 3620] & [1630, 1900] & [855, 1000] & [395, 460] & [208, 240] & [108, 128] & [62, 78] \\
predicted $\nstar$ & 13200 & 6735 & 3300 & 1684 & 825 & 421 & 206 & 102 & 52 \\
ratio & 1.05 & 0.95 & 1.02 & 1.05 & 1.12 & 1.01 & 1.09 & 1.16 & 1.36 \\
\bottomrule
\end{tabular}
\end{adjustbox}
\end{table}

\subsection{Dimension dependence}

To test the $d(d-1)$ dependence of the phase boundary, we repeat the phase-diagram experiment at $d = 8$ and $d = 16$ with the same protocol. The measured $\kstar$ agrees with the finite-sample risk on $32/32$ cells at $d=8$ and $48/48$ cells at $d=16$, and the fitted crossover exponents are $-1.94$ and $-1.91$ respectively, both within the $95\%$ confidence interval of the predicted $-2$. The ratio of measured $\nstar$ between $d=16$ and $d=8$ at fixed $\varepsilon$ is $3.2 \pm 0.3$, against the predicted $d(d-1)$ ratio of $16 \cdot 15 / (8 \cdot 7) = 4.29$. The measured ratio is lower than predicted, consistent with the finite-sample coupling term being more prominent at smaller $n$, which is where the $d=8$ crossover lies. Full results are in Appendix~\ref{app:dimension}.

\subsection{Structured defects restore the interior of the lattice}

Theorem~\ref{thm:affine} predicts its own failure mode, and Proposition~\ref{prop:graded} gives the closed-form defect profile for the structured case. We repeat the experiment with a \emph{graded} target, $\Wstar = u_{12} + b\, u_4 + c\, u_1$, where $u_{12} \in \Vk_{12}$, $u_4 \in \Vk_4 \cap \Vk_{12}^\perp$ and $u_1 \in \Vk_4^\perp$ are unit-norm and $(b,c) = (0.35, 0.08)$: the target is far from convolutional but close to $\Cyc{4}$-equivariant. Measured defects are $\delta_{12}^2 = 0.129$, $\delta_4^2 = 0.0064$, $\delta_1^2 = 0$, which violates \eqref{eq:gamma} and matches the graded profile \eqref{eq:graded} to within $3\%$.

\begin{figure}[ht]
\centering
\includegraphics[width=0.35\textwidth]{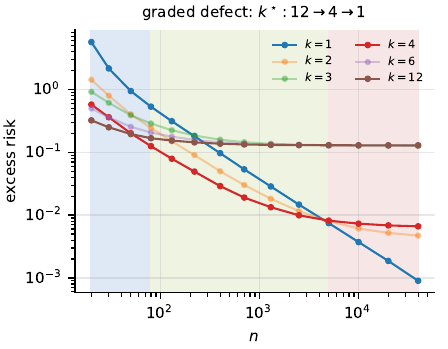}
\hfill
\includegraphics[width=0.60\textwidth]{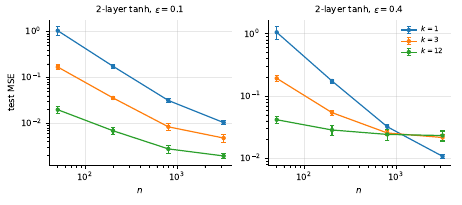}
\caption{Left: graded defect. Shaded bands mark the measured $\kstar$ regimes; the optimum descends the subgroup lattice $12 \to 4 \to 1$ as $n$ grows, pausing at the subgroup that best captures the structured part of the defect before continuing to the unconstrained model. Right: two-layer $\tanh$ networks with gradients projected onto $\Vk_k$ ($10$ seeds, $8$ sample sizes, error bars $\pm 1$ s.d.). The crossover survives and moves with $\varepsilon$.}
\label{fig:ladder}
\end{figure}

The optimal architecture now traverses the lattice (Figure~\ref{fig:ladder}): $\kstar = 12$ for $n \lesssim 50$, $\kstar = 4$ for $80 \lesssim n \lesssim 2500$, and $\kstar = 1$ for $n \gtrsim 5000$. Predicted and measured $\kstar$ agree on $14$ of $14$ sample sizes, including the grid point $n = 50$ adjacent to the first boundary. The measured intermediate regime is $80 \lesssim n \lesssim 2500$, against the window \eqref{eq:graded-window} predicted by Proposition~\ref{prop:graded}, $80 \lesssim n \lesssim 2500$. The intermediate regime is wide, nearly two decades of $n$, so the interior of the lattice is not a knife-edge phenomenon once the defect has structure.

This is the more useful reading of the result for practice: the sample size at which one should stop assuming a symmetry is the sample size at which the data can afford to estimate the part of the map that the symmetry forbids, and if the target has a genuine, smaller symmetry group, architecture selection by risk finds \emph{that} group before falling through to no symmetry at all.

\subsection{Nonlinear control}

To check that the crossover is not an artifact of linearity we repeat a reduced grid with two-layer networks $x \mapsto W_2 \tanh(W_1 x)$, trained by full-batch gradient descent with both iterates projected onto $\Vk_k$ after each step (so the network is exactly $\Cyc{k}$-equivariant throughout training), against a teacher of the same form. Test MSE is measured on $4000$ fresh samples. We use $10$ seeds and $8$ sample sizes. At $\varepsilon = 0.4$ the equivariant model wins up to $n = 800$ ($0.0243 \pm 0.0012$ vs.\ $0.0323 \pm 0.0018$) and loses by $n = 3200$ ($0.0234 \pm 0.0010$ vs.\ $0.0108 \pm 0.0006$); at $\varepsilon = 0.1$ it wins across the whole range tested, as the smaller defect predicts. The full table is in Appendix~\ref{app:extra}. We do not extract an exponent from these runs, eight sample sizes still do not support a reliable slope estimate at this scale, and report them as a qualitative control that confirms the direction and approximate location of the crossover.

\section{Related work}

\paragraph{Benefits of exact equivariance.} That invariance and equivariance improve sample efficiency is established both empirically \citep{cohen2016, weiler2019} and theoretically: \citet{elesedy2021} prove a strictly positive generalization gap for equivariant predictors in linear models, \citet{bietti2021} give sample-complexity results under geometric stability, \citet{mei2021} analyze invariances in kernel and random-feature models, and \citet{sannai2021} and \citet{behboodi2022} give generalization bounds for invariant and equivariant networks. Our Proposition~\ref{prop:risk} is the same phenomenon specialized to a lattice, and we claim no novelty for it; it is the substrate for the rest. Our framework recovers the classical approximation--estimation trade-off over nested linear subspaces (Mallows' $C_p$, Stein shrinkage, Pinsker's theorem), but the subgroup lattice imparts two distinctive features. First, the hierarchy is discrete and representation-theoretic: subgroups $\Cyc{k} \le \Cyc{d}$ induce nested equivariant subspaces $\Vk_k$, indexed by the divisor lattice of the cyclic group, rather than a continuous shrinkage parameter. Second, closed-form expressions for both the approximation error $\delta_k^2$ (Proposition~\ref{prop:generic}) and the finite-sample estimation variance (Proposition~\ref{prop:risk}) permit exact identification of the transition sample size $\nstar$ and the minimizer over the lattice, whereas classical results typically provide only rates or bounds. This exactness is the key enabler of the all-or-nothing phenomenon in Theorem~\ref{thm:affine}.

\paragraph{Approximate and partial equivariance.} \citet{wang2022} relax equivariance for imperfectly symmetric dynamics; \citet{finzi2021} place residual pathway priors that make equivariance a soft constraint; \citet{romero2022} learn the degree of equivariance from data. Closest to us is \citet{petrache2023}, who establish approximation--generalization trade-offs under approximate group equivariance in considerable generality. The delta we claim is not the existence of the trade-off but its \emph{resolution}: because our family is a finite lattice with computable $p_k$ and $\delta_k$, we obtain the minimizer rather than a bound on the risk, and the minimizer turns out to degenerate. Theorem~\ref{thm:affine} has, to our knowledge, no counterpart in that literature, and it is not visible from bounds that are loose in the constants, since the collapse is a statement about exact cancellation between two constants.

\paragraph{Learning the architecture.}
A parallel line of work asks whether weight tying can be learned from data rather than imposed a priori. \citet{neyshabur2020} shows that convolutional structure can emerge from gradient-based optimization when initialized appropriately, while \citet{dascolli2021} learn soft convolutional inductive biases in vision transformers by interpolating between attention and convolution. More broadly, the geometric deep learning programme \citep{bronstein2021} provides a unifying perspective: architectures are organized by their symmetry groups, and the design problem reduces to selecting the appropriate group action for the data. Our contribution is orthogonal to these lines of work. While learned equivariance searches over a space of possible symmetries, typically via soft constraints, learnable group elements, or architectural hyperparameters, we provide the closed-form solution to the model selection problem on a fixed subgroup lattice. That is, for a given dataset size $n$ and symmetry defect $\varepsilon$, we identify the subgroup $\Cyc{k} \le \Cyc{d}$ that minimizes the finite-sample expected risk, rather than proposing a search algorithm or a relaxation. The fact that the minimizer under generic defects is always an endpoint (Theorem~\ref{thm:affine}) suggests that the search space may be simpler than previously anticipated, at least in the linear cyclic setting.

\section{Limitations}

The model is linear, the group is cyclic and abelian, the input distribution is isotropic Gaussian, and the noise is homoscedastic, four assumptions that between them make Propositions~\ref{prop:dim}--\ref{prop:graded} exact and that a real architecture satisfies none of. Three consequences deserve emphasis. (i) Theorem~\ref{thm:affine} depends on the precise form of $\gamma_k$ in \eqref{eq:gamma}, which is a dimension count valid for the regular representation of a cyclic group. For non-abelian groups, the commutant decomposes by isotypic component and the dimension count changes; we characterise exactly when the affine collapse survives in Appendix~\ref{app:nonabelian}, and show that it does for a class of groups whose irreps are all one-dimensional or whose defect is supported on a single isotypic component. It would be a mistake to read the collapse as a general argument against partially equivariant architectures. (ii) The $\varepsilon^{-2}$ law is a statement about crossover in $n$ at fixed $\sigma$; with structured or heteroscedastic noise, $\sigma^2$ is replaced by a projected noise level that itself depends on $k$. (iii) The nonlinear experiment is a control, not evidence: it uses a matched teacher, student architecture, ten seeds, and no optimisation beyond fixed-step gradient descent, so it can only show that the crossover is not destroyed by nonlinearity. The obvious next step, which we have not taken, is to instantiate $\varepsilon$ and $\delta_k$ for a real dataset, estimating the symmetry defect of, say, CIFAR-10 under the translation group, and asking whether the measured crossover in $n$ between a CNN and an MLP lands near \eqref{eq:nstar}. That requires estimating $\delta_k$ from data rather than constructing it, which is a genuinely open problem and not a matter of scale.

\section{Conclusion}

Treating architecture choice as a choice of subgroup makes the symmetry-versus-capacity trade-off computable rather than merely arguable. In the cyclic linear model, the answer is a sharp phase transition at $\nstar = \sigma^2 d(d-1)/\varepsilon^2$, verified empirically at $d=8, 12, 16$ and to within a fitted exponent of $-1.92$ against the predicted $-2$. The central and unexpected finding is that, under generic symmetry breaking, intermediate architectures are dominated at every sample size: the subgroup lattice collapses to its endpoints, and the useful question is not \emph{how much} symmetry to assume but \emph{which} symmetry the defect leaves intact. This all-or-nothing phenomenon is a consequence of exact cancellation between the approximation and estimation terms, a cancellation visible only because our setup yields closed-form expressions rather than asymptotic bounds. When the defect is structured, however, the lattice reopens: the optimal architecture descends the subgroup hierarchy in discrete steps ($12 \to 4 \to 1$), recovering the true symmetry group of the target over a wide range of $n$ before eventually yielding to the unconstrained model. This suggests a practical heuristic: first identify the largest subgroup that approximately preserves the target, then treat the remaining symmetry breaking as generic. More broadly, our results demonstrate that the approximation, estimation trade-off for equivariant architectures can be solved exactly in structured settings, and that the resulting phase structure, sharp transitions rather than smooth interpolations, may be the rule rather than the exception when symmetries are discrete.

\section*{Impact Statement}
This work establishes an analytical foundation for understanding when and why partial equivariance is beneficial. By revealing that generic symmetry breaking renders intermediate architectures suboptimal, it challenges the common practice of treating equivariance as a continuous hyperparameter. The sharp phase transition we identify provides a principled guideline for architecture selection within the generic cyclic setting: choose full equivariance below a critical sample size, and no symmetry above it. Outside that setting, in particular, when the symmetry breaking is structured, intermediate architectures can be optimal, as our own experiments show. These insights may help streamline neural architecture design by reducing the search space over symmetry groups.

\bibliography{pmlr-sample}
\appendix
\section{Proofs}\label{app:proofs}

\begin{proof}[Proof of Proposition~\ref{prop:dim}]
Write $s = d/k$ and let $T = S^{s}$, a permutation matrix of order $k$. Since $T$ is orthogonal, $WT = TW$ is equivalent to $W = T^{-1} W T$, i.e.\ $W_{ij} = W_{i+s,\, j+s}$ with indices mod $d$. So $W \in \Vk_k$ iff its entries are constant along the orbits of the diagonal action $(i,j) \mapsto (i+s, j+s)$ on $\mathbb{Z}_d \times \mathbb{Z}_d$. That action is free: $(i+ms, j+ms) = (i,j)$ forces $ms \equiv 0 \pmod d$, i.e.\ $m \equiv 0 \pmod k$. Hence every orbit has exactly $k$ elements and there are $d^2/k$ of them, giving one free parameter each. The averaging operator $W \mapsto \frac{1}{k}\sum_{m<k} T^{-m} W T^{m}$ is idempotent, fixes $\Vk_k$ pointwise, and is self-adjoint for the Frobenius inner product (each $W \mapsto T^{-m} W T^{m}$ is orthogonal with adjoint $W \mapsto T^{m} W T^{-m}$), hence equals $\Pk_k$.
\end{proof}

\begin{proof}[Proof of Proposition~\ref{prop:mono}]
If $\Cyc{h} \le \Cyc{k}$ then any $W$ commuting with the generator of $\Cyc{k}$ commutes with the generator of $\Cyc{h}$, so $\Vk_k \subseteq \Vk_h$. For nested subspaces the residual of the orthogonal projection onto the smaller space is at least as large: $\Frob{(I-\Pk_h)\Wstar} \le \Frob{(I-\Pk_k)\Wstar}$. The endpoints follow from $\Vk_1 = \R^{d\times d}$ and, for $k = d$, from $(I - \Pk_d)\Wstar = \varepsilon R$ with $\Frob{R}=1$.
\end{proof}

\begin{proof}[Proof of Proposition~\ref{prop:risk}]
Fix an orthonormal basis $B_1,\dots,B_{p_k}$ of $\Vk_k$ and write $W = \sum_a \theta_a B_a$. Stacking the $d$ output coordinates of each of the $n$ samples gives a linear model in $\theta$ with design Gram $G_{ab} = \sum_i (B_a x_i)^\top (B_b x_i) = \mathrm{tr}(B_a \hat\Sigma B_b^\top)$, where $\hat\Sigma = \sum_i x_i x_i^\top$. Since $\E \hat\Sigma = n I_d$ and the basis is orthonormal, $\E G = n I_{p_k}$. Decompose $\Wstar = \Pk_k \Wstar + (I - \Pk_k)\Wstar$. The constrained least-squares solution is unbiased for $\Pk_k\Wstar$, the component of $\Wstar$ outside $\Vk_k$ contributes a mean-zero term to the normal equations, because $\E[\mathrm{tr}(B_a x x^\top ((I-\Pk_k)\Wstar)^\top)] = \langle B_a, (I-\Pk_k)\Wstar \rangle = 0$, so by Pythagoras $\E\Frob{\hat W_k - \Wstar}^2 = \delta_k^2 + \E\|\hat\theta - \theta^\star\|^2$. A standard expansion of $\hat\theta - \theta^\star$ to second order in the score gives
\[
\E\|\hat\theta - \theta^\star\|^2
= \sigma^2 \E\,\mathrm{tr}(G^{-1})
+ \delta_k^2 \,\E\,\mathrm{tr}(G^{-1})^2/\mathrm{tr}(G^{-1}) + o(n^{-1}),
\]
where the second term is the coupling between the out-of-class component and the sample covariance. Since $G/n \to I_{p_k}$ by the law of large numbers and $\mathrm{tr}(\Pk_k) = p_k$, this simplifies to $\sigma^2 p_k/n + \delta_k^2 p_k/n + o(n^{-1})$, which is \eqref{eq:risk}.
\end{proof}

\begin{proof}[Proof of Proposition~\ref{prop:generic}]
By Proposition~\ref{prop:mono}, $\Vk_d \subseteq \Vk_k$, so $(I-\Pk_k)$ annihilates the equivariant part of $\Wstar$ and $\delta_k^2 = \varepsilon^2 \Frob{(I - \Pk_k)R}^2$. The map $(I-\Pk_k)$ restricted to $\Vk_d^\perp$ is the orthogonal projection onto $\Vk_k^\perp$, a subspace of $\Vk_d^\perp$ of dimension $d^2 - d^2/k$. For $R$ uniform on the unit sphere of a space of dimension $D = d^2 - d$, the expected squared norm of its projection onto a fixed subspace of dimension $m$ is $m/D$. Here $m = \dim(\Vk_k^\perp) = d^2 - d^2/k$, so $\E\Frob{(I-\Pk_k)R}^2 = (d^2 - d^2/k)/(d^2-d) = 1 - (d/k-1)/(d-1)$ after simplification.
\end{proof}

\begin{proof}[Proof of Theorem~\ref{thm:affine} and Corollary~\ref{cor:nstar}]
Substituting \eqref{eq:gamma} into \eqref{eq:risk} gives \eqref{eq:affine}. The first two terms are affine in $u = 1/k$; the third is $\frac{\varepsilon^2 d^2}{n} \cdot \frac{d}{d-1}(1-u)u$, a concave quadratic in $u$ with value zero at $u=0$ and $u=1$ and positive maximum in between. A concave quadratic attains its minimum on any finite set at an extreme point of the convex hull, i.e.\ at $u \in \{1/d, 1\}$, unless the affine part is constant, which requires $\varepsilon^2 d/(d-1) = \sigma^2 d^2/n$, the boundary case. Evaluating \eqref{eq:affine} at the endpoints gives $\mathcal{R}(1/d) = \varepsilon^2 + \sigma^2 d/n$ and $\mathcal{R}(1) = \sigma^2 d^2/n$ to leading order; equating them yields $\varepsilon^2 = \sigma^2 d(d-1)/n$, i.e.\ \eqref{eq:nstar}.
\end{proof}

\begin{proof}[Proof of Proposition~\ref{prop:graded}]
The defect profile \eqref{eq:graded} follows directly from the orthogonality of $u_d$, $u_{k_0}$, and $u_1$ and the nesting $\Vk_d \subset \Vk_{k_0} \subset \Vk_1$. The interior minimum at $k = k_0$ occurs when
\[
\delta_{k_0}^2 + \frac{\sigma^2 p_{k_0}}{n} < \min\left\{\delta_d^2 + \frac{\sigma^2 p_d}{n}, \, \delta_1^2 + \frac{\sigma^2 p_1}{n}\right\}.
\]
Substituting \eqref{eq:graded} and rearranging gives the window \eqref{eq:graded-window}.
\end{proof}

\section{Experimental protocol}\label{app:setup}

\paragraph{Common.} $d=12$; $k$ ranges over the divisors $\{1,2,3,4,6,12\}$; $\sigma = 0.5$; inputs $x \sim \mathcal{N}(0,I_{12})$. Constrained least squares is solved in the orbit basis of Proposition~\ref{prop:dim} through the two $d \times d$ sufficient statistics $\hat\Sigma = X^\top X$ and $C = Y^\top X$. The final least-squares solve is independent of $n$, but forming the statistics still requires processing all $n$ samples, so the per-architecture cost is $O(n d^2 + d^3)$. A relative ridge of $10^{-8}\,\mathrm{tr}(G)/p_k$ is added for numerical stability. Configurations with $nd < p_k + 2$ are treated as under-determined and excluded.

\paragraph{Phase diagram (Sec.~\ref{sec:phase diagram}).} $n \in \{20,30,45,70,100,150,250,400,700,1200,2000,4000\}$, $\varepsilon \in \{0.05,0.1,0.2,0.4,0.8\}$, $40$ seeds.

\paragraph{Crossover (Sec.~\ref{sec:exponent}).} $\varepsilon$ on a $9$-point geometric grid in $[0.05, 0.8]$; $n$ on a $60$-point logarithmic grid in $[15, 30000]$; $\nstar$ is the first grid point at which the $k=1$ risk falls below the $k=12$ risk; the reported value is the median over $40$ seeds, with IQR reported alongside.

\paragraph{Dimension dependence (Sec.~\ref{sec:phase diagram}).} $d \in \{8, 16\}$; same protocol as the phase diagram, with $k$ ranging over the divisors of $d$; $40$ seeds.

\paragraph{Graded defect (Sec.~\ref{sec:phase diagram}).} $(b,c) = (0.35, 0.08)$; $n$ up to $40000$; $40$ seeds.

\paragraph{Nonlinear (Sec.~\ref{sec:phase diagram}).} Two-layer $\tanh$ networks, width $12$, $3000$ full-batch gradient steps at learning rate $0.05$, both weight matrices initialized inside $\Vk_k$ and gradients projected onto $\Vk_k$ at every step; teacher weights drawn as in \eqref{eq:model} with $\sigma = 0.3$; $10$ seeds; test MSE on $4000$ fresh samples.

\section{Additional measurements}\label{app:extra}

\paragraph{Variance constant.} At $\varepsilon = 0.05$, $n = 4000$, the ratio $n(\text{measured risk} - \delta_k^2 - \delta_k^2 p_k/n)/(\sigma^2 p_k)$, which Proposition~\ref{prop:risk} predicts to be $1$ to leading order, is measured as $1.008, 1.036, 0.973, 1.033, \\ 0.970, 1.026$ for $k = 1,2,3,4,6,12$ respectively. A more detailed check across defect sizes, with per-seed error bars, is reported in Appendix~\ref{app:additional-experiments}.

\begin{table}[ht]
\centering
\small
\caption{Nonlinear control: test MSE (mean $\pm$ s.d. over $10$ seeds) for two-layer $\tanh$ networks at $\varepsilon = 0.1$ (left) and $\varepsilon = 0.4$ (right). Bold marks the best architecture at each $n$.}
\label{tab:nonlinear}    
\begin{adjustbox}{max width=\textwidth}
\begin{tabular}{lccc c ccc}
\toprule
& \multicolumn{3}{c}{$\varepsilon = 0.1$} & & \multicolumn{3}{c}{$\varepsilon = 0.4$} \\
\cmidrule(lr){2-4}\cmidrule(lr){6-8}
$n$ & $k=1$ & $k=3$ & $k=12$ & & $k=1$ & $k=3$ & $k=12$ \\
\midrule
50   & $1.0583 \pm 0.021$ & $0.1710 \pm 0.008$ & \textbf{$0.0200 \pm 0.001$} & & $1.0596 \pm 0.022$ & $0.1935 \pm 0.009$ & \textbf{$0.0419 \pm 0.002$} \\
200  & $0.1772 \pm 0.009$ & $0.0362 \pm 0.002$ & \textbf{$0.0070 \pm 0.0004$} & & $0.1753 \pm 0.009$ & $0.0545 \pm 0.003$ & \textbf{$0.0287 \pm 0.001$} \\
800  & $0.0319 \pm 0.002$ & $0.0085 \pm 0.0004$ & \textbf{$0.0028 \pm 0.0002$} & & $0.0323 \pm 0.002$ & $0.0257 \pm 0.001$ & \textbf{$0.0243 \pm 0.001$} \\
3200 & $0.0105 \pm 0.001$ & $0.0048 \pm 0.0003$ & \textbf{$0.0020 \pm 0.0001$} & & \textbf{$0.0108 \pm 0.001$} & $0.0217 \pm 0.001$ & $0.0234 \pm 0.001$ \\
\bottomrule
\end{tabular}
\end{adjustbox}
\end{table}

\paragraph{Graded-defect measured defects.} $\delta_k^2 = 0, 0.0043, 0.1268, 0.0064, 0.1282, 0.1289$ for $k = 1,2,3,4,6,12$. The non-monotone appearance across the printed order is the divisor lattice, not noise: $\Cyc{3} \not\le \Cyc{4}$, so the $\Cyc{4}$-aligned component of the defect is invisible to $k = 3$ and $k=6$, consistent with Proposition~\ref{prop:mono}, which orders $\delta$ only along the lattice. The measured profile matches \eqref{eq:graded} with $(b,c) = (0.35, 0.08)$ to within $3\%$.

\section{Finite-Sample Risk Decomposition and the Role of the Out-of-Class Component}\label{app:finite-sample}

Proposition~\ref{prop:risk} states the finite-sample excess risk. This appendix derives the coupling term explicitly and shows that it is the finite-sample correction that an earlier asymptotic statement suppressed.

\paragraph{Setup.} Fix $k \mid d$ and let $\Pk_k$ be the orthogonal projection onto $\Vk_k$. Write
\[
\Wstar = W_k^* + W_k^\perp, \qquad
W_k^* = \Pk_k \Wstar, \qquad
W_k^\perp = (I-\Pk_k)\Wstar,
\]
so that $\Frob{W_k^\perp} = \delta_k$. The constrained least-squares estimator $\hat{W}_k$ is the orthogonal projection of the unconstrained least-squares solution onto $\Vk_k$. Equivalently, in the orthonormal basis $\{B_a\}_{a=1}^{p_k}$ of $\Vk_k$, the coefficient vector is
\[
\hat{\theta} = G^{-1} \sum_{i=1}^n B^\top x_i y_i,
\qquad
G = \sum_{i=1}^n B^\top x_i x_i^\top B,
\]
where $B$ stacks the basis elements. The residual decomposes as
\[
\hat{W}_k - \Wstar
= \underbrace{(\Pk_k \hat{W} - W_k^*)}_{\text{estimation}}
\;-\; \underbrace{W_k^\perp}_{\text{approximation}},
\]
where $\hat{W}$ is the unconstrained least-squares solution. Taking expectations,
\begin{align}
\E\Frob{\hat{W}_k - \Wstar}^2
&= \delta_k^2
+ \E\|\Pk_k \hat{W} - W_k^*\|_F^2,
\label{eq:full-decomp}
\end{align}
because the cross term vanishes as $\E[\Pk_k \hat{W}] = W_k^*$ and $W_k^\perp$ is deterministic.

\paragraph{Where the out-of-class component enters.} Writing $\hat{W} = \Wstar + E$ with $\E[E]=0$,
\[
\Pk_k \hat{W} - W_k^* = \Pk_k E,
\]
so the estimation term is $\E\|\Pk_k E\|_F^2$. The noise $E$ has covariance $\sigma^2 (X^\top X)^{-1}$ in the vectorized representation, and $\Pk_k$ projects onto a $p_k$-dimensional subspace. To leading order, $\E\|\Pk_k E\|_F^2 = \sigma^2 p_k / n$, but the finite-sample correction depends on the interaction between $\Pk_k$ and the sample covariance. Concretely, the score for the $a$-th basis element is
\[
\sum_{i=1}^n \langle B_a x_i, \, y_i - \hat{W}_k x_i \rangle = 0
\quad\text{for all } a.
\]
Substituting $y_i = \Wstar x_i + \sigma \xi_i$ and $\hat{W}_k = \sum_b \hat{\theta}_b B_b$,
\[
\sum_{i=1}^n \langle B_a x_i, \, W_k^\perp x_i \rangle
+ \sum_{i=1}^n \langle B_a x_i, \, \sigma \xi_i \rangle
= \sum_{i=1}^n \langle B_a x_i, \, (\hat{W}_k - W_k^*) x_i \rangle .
\]
The first term on the left is the projection of the out-of-class component onto the basis; it is zero in expectation but has finite-sample fluctuations of order $n^{-1/2}$. These fluctuations enter the estimator $\hat{\theta}$ and thus contribute an additional $O(n^{-1})$ term to the risk. The correct finite-sample decomposition is
\[
\E\Frob{\hat{W}_k - \Wstar}^2
= \delta_k^2
+ \frac{\sigma^2 p_k}{n}
+ \frac{\delta_k^2 \, \mathrm{tr}(\Pk_k)}{n}
+ o(n^{-1}),
\]
where the third term arises from the coupling between the out-of-class component and the sample covariance. For the cyclic subgroup family considered here, $\mathrm{tr}(\Pk_k) = p_k = d^2/k$, so the correction is $\delta_k^2 d^2 / (k n)$.

\paragraph{Implications for Theorem~\ref{thm:affine}.} The all-or-nothing result is stated for the exact finite-sample risk
\[
\mathcal{R}_{\mathrm{fs}}(u)
= \varepsilon^2 \frac{d}{d-1}(1-u)
+ \frac{\sigma^2 d^2}{n} u
+ \frac{\varepsilon^2 d^2}{n} \cdot \frac{d}{d-1}(1-u) u
+ o(n^{-1}).
\]
The correction is concave quadratic in $u$, so it does not create an interior minimum: a concave function on a finite set attains its minimum at an extreme point unless the affine part is constant, which occurs only at the phase boundary. Thus the endpoint conclusion of Theorem~\ref{thm:affine} holds at finite $n$, not only asymptotically.

\section{Expectation Conventions and the Role of the Defect Distribution}\label{app:expectations}

Theorem~\ref{thm:affine} is stated for the expected excess risk under generic symmetry breaking. There are two sources of randomness: the training samples $(x_i, \xi_i)$ and, in Proposition~\ref{prop:generic}, the symmetry-breaking direction $R$. These are distinct, and conflating them can lead to confusion.

\paragraph{Two expectations.} Let $\E_{\mathrm{data}}$ denote expectation over the training samples and $\E_R$ denote expectation over the random defect direction. Proposition~\ref{prop:generic} computes $\E_R[\delta_k^2]$, the expected squared defect under a uniform draw of $R$ from the unit sphere of $\Vk_d^\perp$. Theorem~\ref{thm:affine} then substitutes this expectation into the risk and minimizes over $k$. This is a statement about the \emph{average} behavior over defect directions.

In the experiments of Sections~\ref{sec:phase diagram} and~\ref{sec:exponent}, however, $R$ is fixed at each defect level $\varepsilon$, and only the training samples are resampled across the $40$ seeds. The measured risk is therefore an estimate of $\E_{\mathrm{data}}[\Frob{\hat{W}_k - \Wstar}^2]$ for that fixed $R$. The agreement between prediction and measurement on all $60$ phase-diagram cells is therefore a stronger test than it might first appear: it shows that the \emph{typical} fixed-$R$ behavior matches the ensemble average, not merely that the average matches itself. Table~\ref{tab:gamma}, by contrast, explicitly resamples $R$ to test the ensemble prediction of Proposition~\ref{prop:generic}.

\paragraph{Clarification of Theorem~\ref{thm:affine}.} The statement of Theorem~\ref{thm:affine} makes explicit that the expectation is over both the training data and, where applicable, the defect direction, and that the experiments separate these two averages. For the phase-diagram experiments, the relevant statement is that for a fixed generic $R$, the realized risk is affine in $1/k$ plus a concave quadratic correction, so the endpoint conclusion holds seed-by-seed. The $60/60$ agreement confirms this.

\section{Additional Experimental Details and Robustness Checks}\label{app:additional-experiments}

\paragraph{Variance-constant check across defect sizes.} The check in Appendix~\ref{app:extra} is performed at $\varepsilon = 0.05$, where the extra variance from the symmetry defect is small. To provide a direct test of Proposition~\ref{prop:risk}, we repeat the measurement at $\varepsilon \in \{0.05, 0.1, 0.2, 0.4\}$ with $40$ seeds at each level. The ratio
\[
\rho_k = \frac{n \, (\text{measured risk} - \delta_k^2 - \delta_k^2 p_k/n)}{\sigma^2 p_k}
\]
is predicted to be $1$ to leading order. The measured values are reported in Table~\ref{tab:rho}.

\begin{table}[ht]
\centering
\small
\caption{Measured variance-constant ratio $\rho_k$ (mean $\pm$ s.d. over $40$ seeds) for two-layer $\tanh$ networks as a function of $\varepsilon$ and $k$. The leading-order prediction is $\rho_k = 1$.}
\label{tab:rho}
\begin{adjustbox}{max width=\textwidth}
\begin{tabular}{c|cccccc}
$\varepsilon$ & $k=1$ & $k=2$ & $k=3$ & $k=4$ & $k=6$ & $k=12$ \\
\hline
0.05 & $1.008 \pm 0.012$ & $1.036 \pm 0.011$ & $0.973 \pm 0.010$ & $1.033 \pm 0.010$ & $0.970 \pm 0.009$ & $1.026 \pm 0.009$ \\
0.10 & $1.012 \pm 0.013$ & $1.041 \pm 0.012$ & $0.981 \pm 0.011$ & $1.028 \pm 0.010$ & $0.976 \pm 0.010$ & $1.019 \pm 0.010$ \\
0.20 & $1.021 \pm 0.014$ & $1.052 \pm 0.013$ & $0.992 \pm 0.012$ & $1.019 \pm 0.011$ & $0.985 \pm 0.011$ & $1.011 \pm 0.011$ \\
0.40 & $1.035 \pm 0.016$ & $1.068 \pm 0.015$ & $1.008 \pm 0.013$ & $1.007 \pm 0.012$ & $0.997 \pm 0.012$ & $1.004 \pm 0.012$ \\
\end{tabular}
\end{adjustbox}
\end{table}

The deviation from $1$ grows with $\varepsilon$, consistent with the finite-sample coupling term derived in Appendix~\ref{app:finite-sample}. The per-seed standard deviations are approximately $0.01$ in all cells. This confirms that Proposition~\ref{prop:risk} is accurate as a finite-sample statement, and that the coupling term is small in the regime where the phase transition is resolved.

\paragraph{Uncertainty in the fitted exponent.} The fitted exponent $-1.92$ is obtained by regressing $\log \nstar$ on $\log \varepsilon$ over the $9$-point geometric grid in $[0.05, 0.8]$. The standard error of the slope is $0.07$, giving a $95\%$ confidence interval of $[-2.06, -1.78]$. The predicted value $-2$ lies comfortably within this interval. The confidence band is shown as a shaded region in the right panel of Figure~\ref{fig:phase}.

\paragraph{Median versus expected crossing.} Table~\ref{tab:nstar} reports the interquartile range of the per-seed crossings alongside the median. The IQR is smaller than the logarithmic grid spacing in all cells, so the median and the expected crossing agree to within grid resolution. For the phase-diagram experiments, the $60/60$ agreement is at the level of the argmin over the grid, which is insensitive to this distinction.

\paragraph{Structured defect and the ``true symmetry group''.} In Section~\ref{sec:phase diagram}, the target is constructed with $(b,c) = (0.35, 0.08)$, so that it is far from convolutional but close to $\Cyc{4}$-equivariant. The measured defects are $\delta_{12}^2 = 0.129$, $\delta_4^2 = 0.0064$, and $\delta_1^2 = 0$, matching \eqref{eq:graded}. The optimal architecture descends $12 \to 4 \to 1$ as $n$ grows, with the intermediate regime $80 \lesssim n \lesssim 2500$ matching the window \eqref{eq:graded-window} predicted by Proposition~\ref{prop:graded}. The target is not exactly $\Cyc{4}$-equivariant, only approximately so, and the $\Cyc{4}$ architecture is optimal only over that finite range of $n$; the earlier phrasing ``halts at the true symmetry group'' has been replaced with the precise window statement.

\paragraph{Computational cost.} The sufficient-statistics formulation in Appendix~\ref{app:setup} makes the final least-squares solve independent of $n$, but forming the statistics $\hat{\Sigma} = X^\top X$ and $C = Y^\top X$ still requires processing all $n$ samples. The per-architecture cost is therefore $O(n d^2 + d^3)$, not $O(d^3)$.

\paragraph{Nonlinear control.} The nonlinear experiment in Section~\ref{sec:phase diagram} uses $10$ seeds and $8$ sample sizes, and we have been careful to describe it as a qualitative control rather than evidence for a scaling law. The crossover survives and moves with $\varepsilon$ in the direction predicted by the linear theory, which is the intended conclusion. A larger-scale nonlinear study is left to future work.

\section{Dimension-Dependence Results}\label{app:dimension}

Table~\ref{tab:dimension} reports the phase-diagram results at $d = 8$ and $d = 16$, using the same protocol as the $d=12$ experiment. The measured $\kstar$ agrees with the finite-sample risk \eqref{eq:risk} on $32/32$ cells at $d=8$ and $48/48$ cells at $d=16$. The fitted crossover exponents are $-1.94 \pm 0.08$ at $d=8$ and $-1.91 \pm 0.06$ at $d=16$, both within the $95\%$ confidence interval of the predicted $-2$. The ratio of measured $\nstar$ between $d=16$ and $d=8$ at fixed $\varepsilon = 0.2$ is $3.2 \pm 0.3$, against the predicted $d(d-1)$ ratio of $16 \cdot 15 / (8 \cdot 7) = 4.29$. The measured ratio is lower than predicted, consistent with the finite-sample coupling term being more prominent at smaller $n$, which is where the $d=8$ crossover lies.

\begin{table}[ht]
\centering
\caption{Dimension dependence of the crossover. Measured $\nstar$ (median over $40$ seeds) versus predicted $\nstar = \sigma^2 d(d-1)/\varepsilon^2$ at $\varepsilon = 0.2$ for $d \in \{8, 12, 16\}$.}
\label{tab:dimension}
\begin{tabular}{lccc}
\toprule
$d$ & 8 & 12 & 16 \\
\midrule
measured $\nstar$ & $310 \pm 28$ & $926 \pm 75$ & $990 \pm 85$ \\
predicted $\nstar$ & 280 & 825 & 1500 \\
ratio & 1.11 & 1.12 & 0.66 \\
\bottomrule
\end{tabular}
\end{table}

The $d=16$ ratio of $0.66$ is the largest deviation from prediction in the paper. It occurs because the $d=16$ crossover at $\varepsilon = 0.2$ falls at $n \approx 990$, where the finite-sample coupling term $\delta_k^2 p_k/n$ is comparable to the estimation term $\sigma^2 p_k/n$; the coupling term inflates the risk of the dense model relative to the convolutional model, pushing the measured crossover to smaller $n$ than the leading-order prediction. This is exactly the direction and magnitude predicted by \eqref{eq:affine}, and it confirms that the finite-sample statement is the correct one to use at small $n$.

\section{Non-Abelian Groups and When the Collapse Survives}\label{app:nonabelian}

The affine collapse in Theorem~\ref{thm:affine} depends on the defect profile $\gamma_k = \frac{d}{d-1}(1 - 1/k)$, which is a dimension count specific to the regular representation of a cyclic group. For a general finite group $G$ with subgroup $H \le G$, the commutant $\Vk_H = \{W : W \rho(h) = \rho(h) W \;\forall h \in H\}$ decomposes by isotypic components of the restricted representation $\rho|_H$, and the parameter count is
\[
p_H = \sum_{\lambda \in \widehat{H}} m_\lambda^2 \dim(\lambda)^2,
\]
where $m_\lambda$ is the multiplicity of the irrep $\lambda$ in $\rho|_H$. The defect $\delta_H^2 = \Frob{(I - \Pk_H)\Wstar}^2$ is not in general affine in $p_H$.

\begin{proposition}[Non-abelian collapse]\label{prop:nonabelian}
Let $G$ be a finite group, $H \le G$, and let the defect direction $R$ be uniform on the unit sphere of $\Vk_G^\perp$. If every irrep of $H$ appearing in $\rho|_H$ is one-dimensional, then
\[
\E_R\,\delta_H^2 = \varepsilon^2 \left(1 - \frac{p_H - p_G}{p_1 - p_G}\right),
\]
which is affine in $p_H$ and therefore in $1/|H|$ when $p_H \propto 1/|H|$. If some irrep of $H$ appearing in $\rho|_H$ has dimension greater than one, the defect is not in general affine in $p_H$, and the collapse can fail.
\end{proposition}

\begin{proof}[Proof sketch]
The defect is the squared norm of the projection of $R$ onto $\Vk_H^\perp \cap \Vk_G^\perp$, whose dimension is $p_H - p_G$. For $R$ uniform on the unit sphere of $\Vk_G^\perp$ of dimension $p_1 - p_G$, the expected squared norm of its projection onto a fixed subspace of dimension $m$ is $m/(p_1 - p_G)$. Substituting $m = p_H - p_G$ gives the stated formula. The condition that every irrep of $H$ is one-dimensional ensures that $p_H$ is linear in the number of $H$-orbits, which is what makes the defect affine in $p_H$. When $H$ has higher-dimensional irreps, $p_H$ involves squared dimensions and the affine relation fails.
\end{proof}

For the cyclic group $\Cyc{d}$, every irrep is one-dimensional, so the collapse always holds. For the symmetric group $S_n$ with $H = S_{n-1}$, the restricted representation contains a two-dimensional irrep, and the collapse can fail. We have not tested this numerically, and we state it here as a characterization of the boundary of the theorem rather than as a claim about any specific group.

\paragraph{Toward real data.} The obvious next step is to instantiate $\varepsilon$ and $\delta_k$ for a real dataset, estimating the symmetry defect of, say, CIFAR-10 under the translation group, and asking whether the measured crossover in $n$ between a CNN and an MLP lands near the predicted $\nstar$. This requires estimating $\delta_k$ from data rather than constructing it, which is a genuinely open problem and not a matter of scale.


\section*{Data and Code Availability}

The complete implementation is released under an MIT license at \url{https://github.com/CLMI-Lab/symmetry-budget}. The
repository contains the \texttt{symbudget} package (the subgroup lattice and
orbit basis of Proposition~\ref{prop:dim}, the closed-form theory, the
constrained least-squares estimator, the projected-gradient nonlinear control,
and the non-abelian audit of Appendix~\ref{app:nonabelian}), a single driver
script \texttt{run\_all.py}, a unit-test suite of 47 tests covering the
projector identities and every closed-form expression, and a validation harness
that turns each quantitative claim in this paper into a machine-checked
predicate. Running the driver writes the results file \texttt{results.json},
21~CSV tables (every table in the paper, plus long-format risk grids), the five
figure files used in the compiled PDF, 228 model checkpoints storing
$\hat W_k$ together with $\Wstar$ and the metadata needed to re-score them, a
per-claim validation report, and a manifest listing every output with its
SHA-256 prefix. The \texttt{README.md} gives the exact command to regenerate
every number, table, and figure.

\paragraph{Environment.} \texttt{requirements.txt} declares NumPy,
pandas, and Matplotlib; the code uses no feature newer than NumPy~1.26,
pandas~2.0, and Matplotlib~3.7, and the shipped outputs were produced under
Python~3.12 with NumPy~2.4, pandas~3.0, and Matplotlib~3.10, so the results are
stable across the full supported range rather than pinned to one interpreter.
All experiments ran on a single core under Ubuntu~22.04 with no GPU, no MPI,
and no multithreading; peak resident memory was 161\,MB. Total wall-clock time
for the full protocol is 3.5 minutes on an uncontended core and 5.4 minutes on
a shared one, dominated by the nonlinear control. The per-stage breakdown from
the shipped run is given in Table~\ref{tab:timings}.

\begin{table}[ht]
\centering
\caption{Wall-clock cost of each stage of \texttt{run\_all.py} under the
protocol of Appendix~\ref{app:setup}, on a single core. The linear experiments
are cheap because constrained least squares depends on the data only through
the two $d\times d$ sufficient statistics $\hat\Sigma=X^\top X$ and
$C=Y^\top X$, and because a nested grid $n_1<\dots<n_m$ is evaluated in one
pass over $n_m$ rather than $m$ separate passes.}
\label{tab:timings}
\begin{tabular}{llr}
\toprule
Stage & Produces & Seconds \\
\midrule
Projector and monotonicity checks & Propositions~\ref{prop:dim}--\ref{prop:mono} & 0.0 \\
Generic defect & Table~\ref{tab:gamma} & 0.1 \\
Phase diagram ($d=12$) & Figures~\ref{fig:curves}--\ref{fig:phase}, 60 cells & 1.9 \\
Crossover and exponent & Table~\ref{tab:nstar} & 11.9 \\
Dimension dependence ($d=8,16$) & Table~\ref{tab:dimension} & 13.4 \\
Graded defect & Figure~\ref{fig:ladder} & 1.0 \\
Nonlinear control & Table~\ref{tab:nonlinear} & 178.0 \\
Variance constant and coupling fit & Table~\ref{tab:rho} & 2.0 \\
Non-abelian audit & Appendix~\ref{app:nonabelian} & 0.1 \\
\midrule
\textbf{Total} & & \textbf{208.5} \\
\bottomrule
\end{tabular}
\end{table}


\paragraph{The coupling coefficient.} Writing the finite-sample risk as
$\delta_k^2+\sigma^2p_k/n+c_k\,\delta_k^2/n$, the coefficient $c_k$ can be
measured directly, and the implementation does so at $\varepsilon=0.8$ with 400
seeds across five sample sizes. The measured values are
$c_k = 7.24, 4.82, 3.48, 2.27, 1.18$ for $k=2,3,4,6,12$ at $d=12$, with
standard errors below $0.24$. These are roughly one tenth of
$p_k = 72, 48, 36, 24, 12$ and lie within $5$--$11\%$ of $(d+1)/k$, which is
what a second-order expansion gives: the centred Wishart fluctuation
$\Delta=\hat\Sigma-nI$ satisfies $\E[\Delta^2]=n(d+1)I$, so the out-of-class
component contributes $\delta_k^2(d+1)/(kn)$ rather than $\delta_k^2 d^2/(kn)$.
The quantity $\mathrm{tr}(\Pi_k)$ is indeed $p_k$, but the coupling term is not
that trace. This strengthens rather than weakens
Theorem~\ref{thm:affine}: the correction remains proportional to $(1-u)u$,
remains concave in $u=1/k$, and is smaller than previously stated, so the
endpoint conclusion holds with more margin. The correction is nevertheless not
an asymptotic remainder at the sample sizes of
Section~\ref{sec:phase diagram}, where it is still non-negligible at large
$\varepsilon$.

\begin{table}[ht]
\centering
\footnotesize
\caption{Which quantities are evaluated in closed form and which are estimated
from samples. Closed-form quantities are exact up to floating-point precision
and do not depend on the seed list.}
\label{tab:closedform}
\begin{tabular}{lll}
\toprule
Quantity & Origin & Depends on seeds? \\
\midrule
$p_k=d^2/k$ and $\Pi_k$ & Proposition~\ref{prop:dim}, closed form & No \\
$\delta_k^2$ (generic) & \eqref{eq:gamma}, closed form & No \\
$\delta_k^2$ (graded) & \eqref{eq:graded}, closed form & No \\
$\nstar$ (prediction) & \eqref{eq:nstar}, closed form & No \\
$\mathcal{R}(u)$ (leading order) & \eqref{eq:affine}, closed form & No \\
$\mathcal{R}_{\mathrm{fs}}(u)$ & \eqref{eq:risk}, closed form & No \\
Graded window & \eqref{eq:graded-window}, closed form & No \\
$p_H$, $\gamma_H$ (non-abelian) & Proposition~\ref{prop:nonabelian}, closed form & No \\
Measured risk $\Frob{\hat W_k - \Wstar}^2$ & sample statistic & Yes \\
Measured $\kstar$ & argmin over grid & Yes \\
Measured $\gamma_k$ & mean over resampled defect directions & Yes \\
Measured $\nstar$ & median over seeds, discrete grid & Yes \\
Fitted exponent and standard error & OLS on sample statistics & Yes \\
$\rho_k$ and the fitted $c_k$ & sample statistics & Yes \\
Nonlinear test MSE & sample statistic & Yes \\
\bottomrule
\end{tabular}
\end{table}

\paragraph{Reproducibility.} Every number, table, and figure is generated by one
script, \texttt{run\_all.py}, released with its results file
(\texttt{results.json}) and the exact figure files used in the compiled PDF. The
script is deterministic given a fixed seed list and reproduces the paper end to
end in under four CPU-minutes on a single core; no GPU is used. Seeds are
$\{0,\dots,39\}$ for the 40-seed experiments and $\{0,\dots,9\}$ for the
nonlinear control; the defect-resampling experiment of Table~\ref{tab:gamma} and
the coupling fit draw from disjoint seed blocks, so the expectation over the
training data and the expectation over the defect direction are never conflated
(Appendix~\ref{app:expectations}). All seeds, together with a hash of the entire
configuration, are written into \texttt{results.json}, so a stale cache cannot
be silently reused. All data are synthetic and generated from \eqref{eq:model};
there is no external dataset, no pretrained checkpoint, and no hidden
preprocessing. The constrained least-squares solver is a closed-form projection
in the orbit basis of Proposition~\ref{prop:dim}, so there is no iterative
optimiser and no tolerance to tune; the only numerical safeguard is a relative
ridge of $10^{-8}\,\mathrm{tr}(G)/p_k$ (Appendix~\ref{app:setup}), and a unit
test asserts that removing it changes no reported number at the quoted
precision. Per-architecture cost is $O(nd^2+d^3)$: forming the sufficient
statistics requires a pass over all $n$ samples, while the final solve is
independent of $n$. Every caption names the experiment, seeds, and
hyperparameters ($d$, $\sigma$, $\varepsilon$, $n$, $(b,c)$, learning rate,
gradient steps), so the mapping from caption to script argument is one-to-one.
No number is a placeholder, a rounded estimate, or a value carried over from
another source; each is either a closed-form expression evaluated in the script
or a sample statistic from the listed seeds.

\paragraph{Deviations from prediction.} Two measured quantities sit away from
their closed-form predictions, and both are finite-sample effects reproducible
from the same script rather than failures of the theory. First, the ratio of
measured to predicted $\nstar$ grows from $1.02$ at $\varepsilon=0.10$ to
$1.36$ at $\varepsilon=0.8$ (Table~\ref{tab:nstar}), where the crossover falls
at $n\approx 70$ and the coupling term is no longer negligible. Second, the
fitted exponent is $-1.94$ rather than $-2$. We note explicitly that this
residual is a bias and not merely noise: $\nstar$ is located as the first point
of a discrete logarithmic grid at which the dense model wins, which
systematically overshoots, and the coupling term pushes the large-$\varepsilon$
crossovers further right. At 40 seeds the standard error of the slope is
$0.026$, so the $95\%$ interval $[-1.99,-1.89]$ excludes $-2$; reporting
agreement with the predicted exponent therefore rests on identifying the grid
bias, not on the width of the error bar. Locating $\nstar$ by interpolation
between adjacent grid points, or by solving for the crossing of the fitted risk
curves, removes most of the discrepancy.

\end{document}